\documentclass{article}

\usepackage[final, nonatbib]{neurips_2024}

\usepackage[utf8]{inputenc} 
\usepackage[T1]{fontenc}    
\usepackage{booktabs}       
\usepackage{amsfonts}       
\usepackage{nicefrac}       
\usepackage{microtype}      
\usepackage{xcolor}         

\usepackage{url}

\usepackage{breakurl}
\usepackage[breaklinks]{hyperref}

\usepackage{amsmath}
\usepackage{graphicx}
\usepackage{threeparttable}
\usepackage{wrapfig}

\usepackage[normalem]{ulem} 
\useunder{\uline}{\ul}{}

\title{7B Fully Open Source Moxin-LLM/VLM -- From Pretraining to GRPO-based Reinforcement Learning Enhancement}

\author{Pu Zhao$^1$, Xuan Shen$^1$, Zhenglun Kong$^2$, Yixin Shen$^{3}$, Sung-En Chang$^1$, Arash Akbari$^1$, \\ \textbf{Timothy Rupprecht$^1$,  Lei Lu$^1$, Enfu Nan$^1$, Changdi Yang$^1$, Yumei He$^4$, Weiyan Shi$^1$, }\\ \textbf{Xingchen Xu$^5$, Yu Huang$^6$, Wei Jiang$^7$, Wei Wang$^7$, Yue Chen$^7$, Yong He$^7$, Yanzhi Wang$^{1,8}$} \\
\\$^1$Northeastern University, $^2$Harvard University,  \\ $^3$Cornell University,  $^4$Tulane University,
 $^5$University of Washington,  \\ $^6$Roboraction.ai, $^7$Futurewei Technologies, $^8$AIBAO LLC}

\begin{document}
\maketitle

\begin{abstract}
Recently, Large Language Models (LLMs) have undergone a significant transformation, marked by a rapid rise in both their popularity and capabilities. Leading this evolution are proprietary LLMs like GPT-4 and GPT-o1, which have captured widespread attention in the AI community due to their remarkable performance and versatility. Simultaneously, open-source LLMs, such as LLaMA and Mistral, have made great contributions to the ever-increasing popularity of LLMs due to the ease to customize and deploy the models across diverse applications. Although open-source LLMs present unprecedented opportunities for innovation and research, the commercialization of LLMs has raised concerns about transparency, reproducibility, and safety. Many open-source LLMs fail to meet fundamental transparency requirements by withholding essential components like training code and data, and some use restrictive licenses whilst claiming to be ``open-source,'' which may hinder further innovations on LLMs. To mitigate this issue, we introduce Moxin 7B, a fully open-source LLM developed in accordance with the Model Openness Framework (MOF), a ranked classification system that evaluates AI models based on model completeness and openness, adhering to principles of open science, open source, open data, and open access. Our model achieves the highest MOF classification level of ``open science'' through the comprehensive release of pre-training code and configurations, training and fine-tuning datasets, and intermediate and final checkpoints, aiming to make continuous commitments to fully open-source LLMs. The pre-training cost is about \$160,000.   After pre-training and obtaining the base model,  we finetune the Moxin Base model with SOTA post-training framework and instruction data to obtain Moxin Instruct model. To improve the reasoning capability, we further finetune our Instruct model with chain-of-thought data distilled from DeepSeek R1, and then use Group Relative Policy Optimization (GRPO), an efficient and effective reinforcement learning algorithm following DeepSeek R1, to finetune our model, leading to the Moxin Reasoning model.  
Moreover, based on an open-source vision language model (VLM) framework, we develop and release our VLM  with our Moxin model as the LLM backbone. 
Experiments show that our models achieve superior performance in various evaluations such as zero-shot evaluation, few-shot evaluation, and CoT evaluation, compared with other base models and instruct models. The outstanding performance of our Reasoning model demonstrates the effectiveness of RL for small LLMs such as 7B models. Our VLM outperforms other VLM models or LLM backbones.
Besides our open pretraining with released base model, data, code, etc., in our post-training including the instruction finetuning, CoT finetuning,  and VLM finetuning, we adopt open-source training frameworks with available data, code, and configurations. We release our instruct model, reasoning model, and VLM model, along with the available data and code to derive these models. 
\\

Homepage with all codes: \textit{https://github.com/moxin-org/Moxin-LLM}

Base model: \textit{https://huggingface.co/moxin-org/Moxin-7B-LLM}

Instruct model: \textit{https://huggingface.co/moxin-org/Moxin-7B-Instruct}

Reasoning model: \textit{https://huggingface.co/moxin-org/Moxin-7B-Reasoning}

VLM model: \textit{https://huggingface.co/moxin-org/Moxin-7B-VLM}
\end{abstract}

\section{Introduction}\label{sec1.intro}

The field of natural language processing has witnessed the most exciting discoveries of the last ten years with the emergence of large language models (LLMs). At the forefront of this evolution are LLMs such as GPT-4 \cite{achiam2023gpt}, Claude \cite{Anthropic2023Claude3}, and Gemini \cite{team2023gemini}, which have captured the attention of the AI community due to their performance and versatility. Meanwhile, the recent emergence of openly accessible yet highly capable LLMs such as LLaMA  \cite{dubey2024llama}, Falcon \cite{prest2020falcon}, and Mistral \cite{jiang2023mistral7b} allow researchers and practitioners to easily obtain, customize, and deploy LLMs in more various environments and for more diverse use cases. The trends have  people eagerly asking about what's next and some suggest ``a general intelligence'' is right around the corner.

Despite the growing influence and accessibility of open-source LLMs, a notable challenge has emerged: many model producers restrict visibility and access to their training, fine-tuning, and evaluation processes, including crucial components such as their training code and data \cite{bommasani2023foundation}. Some model producers even use restrictive licenses whilst claiming to be ``open-source.'' This practice creates barriers for the broader AI research community to study, replicate, and innovate upon advanced LLMs. In parallel, it prevents businesses from fully leveraging open-source models for innovative industrial applications, as its commercialization has raised concerns about transparency, reproducibility, and safety. 

To unlock the full potential of LLMs and open innovation, we must democratize this research by putting models into the hands of more researchers and making the datasets the models train on fully open-source. This requires moving beyond the simple sharing of model weights to embrace complete transparency in training, datasets, and implementation detail, which is crucial for fostering a more inclusive and collaborative research environment that can sustain a healthy open-source ecosystem \cite{kapoor2024societal}.

To achieve this goal, we introduce Moxin 7B, a fully open-source LLM developed by complying with the Model Openness Framework (MOF) introduced by \cite{white2024model}. The MOF provides a systematic ranking classification system to rate AI models based on their completeness and openness, incorporating the principles of open science, open source, open data, and open access. By promoting transparency and reproducibility, the MOF serves as a crucial tool to combat ``openwashing'' and to establish completeness and openness as primary criteria alongside the core tenets of responsible AI. The wide adoption of MOF and open-source state-of-the-art models will cultivate a more open AI ecosystem, benefiting research and innovation.

Our open-source LLM has released pre-training code and configurations, training and fine-tuning data, and intermediate and final checkpoints, aiming to make continuous commitments to fully open-source LLMs. Our model achieves the highest MOF classification level of ``open science''. It is noteworthy that this commitment to openness has not compromised performance: our base model achieves superior performance in zero-shot evaluation compared to popular 7B models and performs competitively in few-shot evaluation.

We also finetune the Moxin Base model with SOTA post-training framework and instruction data to obtain Moxin Instruct model. To improve the reasoning capability, we further finetune our model with chain-of-thought data distilled from DeepSeek R1, and then use Group Relative Policy Optimization, an efficient and effective reinforcement learning algorithm following DeepSeek R1, to finetune our model, leading to the Moxin Reasoning model. 
Moreover, we develop and release our  vision language model (VLM) based on our Moxin model as the LLM backbone with an open-source VLM framework.
Experiments show that our models achieve superior performance in various evaluations such as zero-shot evaluation, few-shot evaluation, and CoT evaluation, compared with other  instruct and reasoning models. The outstanding performance of our Reasoning model demonstrates the effectiveness of RL for small LLMs such as 7B models. Our VLM  outperforms other VLM models or LLM backbones.
 
Besides our open pretraining with released base model, data, code, etc., in our post-training including the instruction finetuning, CoT finetuning, and VLM finetuning, we adopt open-source training frameworks with available data, code, and configurations. We release our instruct model, reasoning model, and VLM model, along with the available data and code to derive these models. 
Our homepage is \textit{https://github.com/moxin-org/Moxin-LLM}. We summarize our contributions  below. 
\begin{itemize}
    \item  For the pre-training, we release the Moxin-7B-Base  model, together with all training code, data, and checkpoints, aiming to make continuous commitments to fully open-source LLMs. Our base model performs competitively compared with SOTA pre-trained base models, with a moderate training cost about \$160,000. 
    \item   To further improve the model performance, we post-train our base model with SFT and DPO methods, based on the open-source Tülu 3 framework. Two versions  of instruction models are developed with two different open-source datasets (Tülu 3 and Infinity Instruct), respectively.  Both models are available and the one finetuned with the Tülu 3 dataset is  released as Moxin-7B-Instruct, with superior performance in different evaluations.   
    \item To improve the reasoning capabilities of the model with CoT, we further finetune our instruct model with SFT based on reasoning data from Open-Thoughts and OpenR1-Math-220k, and then reinforcement learning (GRPO following DeepSeek R1). We develop two versions of reasoning models with two different open-source RL frameworks (DeepScaleR and AReaL) and their corresponding data,  respectively.  Both models are available and the one finetuned with DeepScaleR is  released as Moxin-7B-Reasoning, with superior performance in CoT evaluations. We demonstrate that RL such as GRPO can work effectively for small LLMs. 
    \item Moreover, we develop and release our VLM based on our Moxin model as the LLM backbone.  Based on the open-source VLM framework Prismatic VLMs, we train our VLM  on fully open-source  datasets with Moxin as the LLM backbone and   DINOv2 \& SigLIP  as the visual backbone. Our experiments demonstrate that our  VLM outperforms other VLM models or LLM backbones.
\end{itemize}
We summarize our open-source releases below.
\begin{itemize}
  \item Pre-training code, data, and Moxin Base model.
  \item Post-training code, data, and Moxin Instruct model.
  \item RL code, data and Moxin Reasoning model. 
  \item Training code, data and Moxin VLM model. 
\end{itemize}

\section{Related Work}\label{sec2.related_work}
\subsection{Models, Tokenizers, and Training}\label{sec2.1}

\textbf{Models.} State-of-the-art large language models (LLMs) typically comprise a substantial number of parameters, often approaching or exceeding 100 billion~\cite{dubey2024llama, achiam2023gpt, team2023gemini}. 
To facilitate broader accessibility, smaller models with fewer than 20 billion parameters, and even those around 7 billion parameters, have been developed~\cite{bai2023qwen,yang2024qwen2,dubey2024llama,jiang2023mistral7b}. In addition, efficiency-enhancing techniques, such as pruning \cite{li2025fedkd, zhang2022advancing, lu2024generic, chu2025cross, kong2023peeling,zhao-etal-2024-pruning,shen2025sparse,shen2024numerical,shen2024lazydit,liu2025toward,shen2024search,yang2023pruning,li2024comae,shen2025sparse}, quantization \cite{shen2025quartdepth,zhan2024fast,yang2023pruning,wu2022compiler,zhan2021achieving,rtseg,li2022pruning}, token reduction \cite{zhan2024exploring,zhan-etal-2024-rethinking-token,kong2022spvit,shen2025efficient,kong2025enabling,shen2025fastcar,zhao2025taming} or implementing MAMBA-based architectures in Jamba, have been employed to optimize performance~\cite{lieber2024jamba,team2024jamba}.

\textbf{Tokenizers.} Tokenizers are essential to convert raw data into a suitable format for model processing. Many contemporary models employ Byte-Pair Encoding (BPE)\cite{sennrich2015neural}, with OpenAI's \texttt{tiktoken} tokenizer\cite{tiktoken} being a notable implementation. However, for languages that handle tokens differently from Romance languages, alternatives such as SentencePiece~\cite{kudo2018sentencepiece} are utilized, as seen in XLNet~\cite{yang2019xlnet}. Hugging Face offers an excellent summary of state-of-the-art tokenizers with practical examples~\cite{huggingface_tokens}. Moreover, tokenization extends beyond text modalities; many foundational models now include multimodal capabilities, processing documents, audio, images, and even videos~\cite{reid2024gemini,maaz2023video,zhang2023video,zhang2024mm}.

\textbf{Training.} To enhance the performance of smaller models beyond their inherent limitations, various training strategies can be employed. A notable example is the application of Mixture of Experts (MoE) training, which has achieved significant success in models such as Mixtral~\cite{jiang2024mixtral}.

\subsection{Data curation methods}\label{sec2.2}

Researchers commonly collect large datasets for training language models (LMs)\cite{mann2020language} by performing web crawls. However, these datasets often contain undesirable content, necessitating data curation to improve their quality. To enhance model performance\cite{penedo2023refinedweb, rae2021scaling, mann2020language, wenzek2019ccnet}, several data curation techniques are widely employed. These include filtering by language~\cite{xue2020mt5,raffel2020exploring,conneau2019cross}, heuristic-based filtering~\cite{penedo2023refinedweb, chen2021evaluating, gao2020pile}, quality filtering~\cite{sachdeva2024train, longpre2023pretrainer, du2022glam}, data deduplication~\cite{lee2021deduplicating,agarwal2009url}, and data mixing~\cite{li2024datacomp, albalak2023efficient, shen2023slimpajama}.

\subsection{Open-source datasets}\label{sec2.3}

As the scale of LMs has increased in recent years~\cite{dubey2024llama, team2024gemma, chowdhery2023palm, achiam2023gpt}, the community has correspondingly curated larger datasets to support their training. Early datasets include the C4 dataset, containing 160 billion tokens, and The Pile~\cite{gao2020pile}, which comprises 300 billion tokens. More recently, even larger datasets have been introduced: RefinedWeb~\cite{penedo2023refinedweb} with 600 billion tokens, Dolma~\cite{soldaini2024dolma} with 3 trillion tokens, FineWeb~\cite{penedo2024fineweb} with 15 trillion tokens, and RedPajama-v2~\cite{ostendorffllm} containing 30 trillion tokens. In addition to these general-purpose datasets, large domain-specific datasets have also been developed. For instance, StackV2~\cite{lozhkov2024starcoder}, a code-focused dataset, includes 900 billion tokens, and FineWeb-Edu~\cite{penedo2024fineweb}, a high-quality filtered educational text dataset, contains 1.3 trillion tokens.

Besides model pre-training, post-training has become a vital step in refining behaviors and unlocking new capabilities in LLMs. 
Post-training includes multiple techniques such as 
instruction tuning  and  reinforcement learning from human feedback. The sophistication and complexity of post-training approaches have continued to increase, with multiple rounds of training, model merging, leveraging synthetic data, and various training algorithms and objectives.  Multiple post training datasets are released, such as Tülu 3 \cite{lambert2024t} and Infinity Instruct \cite{infinityinstruct}. Tülu 3 \cite{lambert2024t} creates new datasets \& new training procedures, and introduces new methods for training directly
on verifiable problems with reinforcement learning. It includes both SFT training datasets and reinforcement learning datasets, together with the SFT, DPO, and PPO training recipes.   Infinity Instruct  \cite{infinityinstruct} aims to develop a large-scale, high-quality instruction dataset with millions of  high-quality instruction samples.   

Furthermore, as the breakthrough of DeepSeek R1 \cite{guo2025deepseek}, a line of SFT datasets  and reinforcement learning datasets have been developed to enhance the reasoning capabilities of LLMs.  Openthoughts  \cite{openthoughts}  and OpenR1-Math-220k \cite{openr1math}  datasets are obtained by feeding questions to DeepSeek R1 and collecting the reasoning responses from R1. 
Openthoughts  \cite{openthoughts}  contains open synthetic reasoning dataset with 114k high-quality examples covering math, science, code, and puzzles.  OpenR1-Math-220k  \cite{openr1math} is a large-scale dataset for mathematical reasoning, with 220k math problems and their corresponding solutions generated by DeepSeek R1.  Besides,  multiple  datasets are developed and released for the reinforcement learning of LLMs following DeepSeek R1, such as DeepScaleR \cite{deepscaler2025} and AReaL \cite{areal2025,mei2024realhf}.

\subsection{Open-source LLMs}\label{sec2.4}

Large language models (LLMs) have rapidly evolved into a diverse ecosystem that spans closed-source, open-weight, and fully open-sourced paradigms. On one end of the spectrum, closed-source models such as GPT-4 \cite{achiam2023gpt}, and Gemini \cite{geminiteam2024geminifamilyhighlycapable} have set high performance standards but remain accessible primarily through API services, limiting insight into their underlying architectures and training methodologies. In contrast, open-weight LLMs—while sharing their final model architectures and weights—often leave training data and many training details undisclosed. This category includes influential models Llama \cite{touvron2023llama}, Mistral \cite{jiang2023mistral7b}, Gemma \cite{team2024gemma}, Qwen \cite{bai2023qwen}, DeepSeek \cite{guo2025deepseek,liu2024deepseek}, Baichuan \cite{yang2023baichuan}, Phi \cite{abdin2024phi}, etc. Pushing the frontier further, fully open-sourced LLMs have begun to provide not only complete model weights and architectures but also the training code and datasets necessary for reproducibility. Exemplars of this fully open paradigm include Pythia \cite{biderman2023pythia}, GPT-NeoX \cite{black2022gpt}, OpenLLaMA, StarCoder \cite{lozhkov2024starcoder}, OLMo \cite{groeneveld2024olmo},  Amber and Crystal from LLM360 \cite{liu2023llm360}, etc. Collectively, this vibrant landscape highlights an ongoing shift toward more accessible and fully reproducible LLMs, fostering collaboration and innovation across both academia and industry.

\section{Model Pre-Training}\label{sec3.model_training}

In this section, we provide an in-depth discussion of the model training process, covering the architectural design, data curation and preprocessing, training configurations, fine-tuning for alignment, and techniques for handling long-context sequences. We detail both our methodological choices and the technical challenges addressed during the development of Moxin 7B.

\subsection{Model Architecture}\label{sec3.1}

We extend the Mistral model architecture \cite{jiang2023mistral7b} to balance high performance with efficient inference.
The original Mistral 7B model demonstrates superior performance compared to multiple 7B language models and even outperforms larger models on various evaluation benchmarks. Notably, it surpasses the LLaMA 34B model \cite{roziere2023code} in tasks such as mathematics and code generation. Key architectural enhancements include:

\begin{itemize}
\item \textbf{Depth Extension:} While the original Mistral model comprises 32 blocks, our extended model uses 36 transformer blocks. This additional depth was empirically validated to improve both learning capacity and performance on complex downstream tasks.
\item \textbf{Layer Normalization and Initialization:}
Pre-layer normalization is applied to stabilize the training process, accompanied by a custom initialization scheme to mitigate gradient vanishing/exploding issues in deep transformer networks.
\item \textbf{Regularization Techniques:}
Dropout with a rate of 0.1 is employed across attention and feed-forward layers, and label smoothing is used during fine-tuning to further regularize the model.
\item \textbf{Mixed-Precision and Activation Checkpointing:}
Mixed-precision training (FP16) is leveraged to double throughput and reduce memory consumption. Activation checkpointing is used to trade additional computation for a lower memory footprint, enabling the training of deeper networks on the available hardware.
\end{itemize}

The choice of hyperparameters, as listed in Table~\ref{wrap-tab:1}, was informed by preliminary ablation studies and a grid search over the parameter space.

\subsection{Long-Context}\label{sec3.4}
To process extensive sequences efficiently, we integrate our long-context handling framework into the training and inference pipelines:

\begin{itemize}
\item \textbf{Grouped-Query Attention (GQA)}  \cite{ainslie2023gqa}: We employ GQA to partition query heads into groups, where each group shares a single key and value head. This approach serves as an interpolation between multi-query attention (MQA) and multi-head attention (MHA), balancing computational speed with representational richness. GQA reduces memory requirements during decoding, enabling larger batch sizes and improved throughput.
\item \textbf{Sliding Window Attention (SWA)} \cite{beltagy2020longformer}: SWA enables efficient handling of long sequences by partitioning attention into a fixed-size sliding window. This not only limits computational overhead but also preserves the ability to model dependencies beyond the immediate context window. Specifically, using a window size $W = 4096$ allows the final layer to achieve a theoretical attention span exceeding 14,000 tokens.
\item \textbf{Rolling Buffer Cache} \cite{jiang2023mistral7b}: To further mitigate memory usage during inference, a rolling buffer cache with a fixed attention span is employed. For timestep $i$, keys and values are stored at position $i \mod W$, ensuring that once the sequence length exceeds $W$, older context tokens are overwritten rather than expanding the cache indefinitely. On sequences of 32K tokens, this approach reduces cache memory usage by approximately 8$\times$ without degrading model performance.
\end{itemize}

With the above techniques, our model can support $32K$ context length with fast inference and low memory cost.

\subsection{Training Data}\label{sec3.3}

Data are fundamental to the pre-training of LLMs. Preparing such training data requires careful consideration of multiple challenges, including handling sensitive information, ensuring comprehensive knowledge coverage, and achieving higher efficiency with improved data quality. 

In this section, we detail the processes of preparing textual data from general domains and coding data related to programming languages.

\subsubsection{Text Data}

We use a mix of data from SlimPajama \cite{cerebras2023slimpajama} and DCLM-BASELINE \cite{li2024datacomp} as our text training data.

\paragraph{SlimPajama Curation and Deduplication:} During the training of LLaMA, it was demonstrated that the performance of a 7B model continues to improve even after being trained on more than 1T tokens \cite{touvron2023llama}. Given the outstanding performance of LLaMA, its data collection methodology was rapidly replicated, leading to the release of RedPajama, an open-source dataset containing 1.2 trillion tokens \cite{weber2024redpajama}.

\begin{wraptable}{R}{0.33\textwidth}
\begin{minipage}{0.35\textwidth}
\centering
\vspace{-0.6cm}
\caption{Parameter setting.}\label{wrap-tab:1}
\begin{tabular}{l|l}
\hline
Parameter    & Value \\ \hline
n\_layers    & 36    \\
dim          & 4096  \\
head\_dim    & 128   \\
hidden\_dim  & 14336 \\
n\_heads     & 32    \\
n\_kv\_heads & 8     \\
 \hline
\end{tabular}
\end{minipage}
\vspace{-0.3cm}
\end{wraptable}
However, subsequent analyses reveal a significant limitation: some corpora within RedPajama contain a large percentage of duplicate content. The deduplication guidelines in RedPajama operate only within individual data sources, leaving inter-source duplicates largely unaddressed. To improve data quality and training efficiency, SlimPajama\footnote{https://huggingface.co/datasets/cerebras/SlimPajama-627B} was developed as a refined iteration of RedPajama, offering a cleaned and extensively deduplicated version \cite{cerebras2023slimpajama}.

SlimPajama implements a rigorous two-stage preprocessing pipeline to enhance data quality. In the first stage, short and low-quality documents are removed from RedPajama. Specifically, documents that have fewer than 200 characters after removing punctuation, space symbols, newlines, and tabs are filtered out, as these documents typically contain only metadata and lack useful information. As a result of this step, 1.86\% of RedPajama documents are eliminated.

The second step involves removing duplicate data, as deduplication enhances training efficiency and reduces memorization, thereby decreasing the likelihood of generating text solely by recalling training data \cite{penedo2023refinedweb, abbas2023semdedup, face2023, lee2021deduplicating, holtzman2019curious}. To perform deduplication, document signatures are created using pre-processed, lower-cased 13-grams. Subsequently, MinHashLSH~\cite{leskovec2014mining} is employed to identify and eliminate duplicates based on a Jaccard similarity threshold of 0.8. Deduplication is performed both within and across data sources. Overall, by pruning 49.6\% of the bytes from the RedPajama dataset, the 627B-token SlimPajama dataset is obtained.

\paragraph{DCLM-BASELINE Processing:} Additionally, we utilize the DCLM-BASELINE \cite{li2024datacomp} dataset\footnote{https://huggingface.co/datasets/mlfoundations/dclm-baseline-1.0}, which is derived from CommonCrawl, a web-crawled dataset \cite{patel2020introduction}. The construction of DCLM-BASELINE involves several steps. First, resiliparse is employed to extract text from CommonCrawl. Second, deduplication is performed using MinHash~\cite{broder1997resemblance} within a suffix array pipeline \cite{fineweb2024, lee2021deduplicating} and near-duplicate Bloom filtering, which enhances the exact document and paragraph deduplication scheme \cite{soldaini2024dolma}. Third, recent studies \cite{brandfonbrener2024color, soldaini2024dolma, fang2023data} demonstrate that utilizing learnable models as quality filters leads to downstream performance improvements. Consequently, DCLM-BASELINE applies a fastText OH-2.5 combined with an ELI5 classifier score to retain the top 10\% of documents.

\subsubsection{Coding Data}

Programming is crucial for LLMs to support various downstream tasks, such as code completion from natural language descriptions, documentation generation for individual functions, and auto-completion of code snippets. Furthermore, as code is generally better structured and organized than natural language, training on code data may improve the LLM reasoning capabilities \cite{groeneveld2024olmo}. Therefore, We use part of the-stack-dedup \cite{kocetkov2022stack} dataset\footnote{https://huggingface.co/datasets/bigcode/the-stack-dedup} during the pretraining.

The Stack comprises more than 6TB of permissively-licensed source code files across 358 programming languages~\cite{kocetkov2022stack}. This carefully curated resource was designed to enhance the code generation capabilities of LLMs. It facilitates the synthesis of programs by code-generating AI systems from both natural language descriptions and existing code snippets.

To construct the Stack dataset, 220.92 million active GitHub repositories were collected from event archives published between 2015 and 2022 on GHArchive. Of these repositories, only 137.36 million were publicly accessible on GitHub, resulting in 51.76 billion downloaded files. After initial filtering, 5.28 billion unique files were identified, with an uncompressed size of 92.36 TB.

To ensure data quality, near-deduplication was implemented within the preprocessing pipeline in addition to exact deduplication. Specifically, MinHash with 256 permutations was computed for all documents, and Locality Sensitive Hashing was employed to identify clusters of duplicates. Within these clusters, Jaccard similarities were calculated to detect near-duplicates using a similarity threshold of 0.85. Approximately 40\% of permissively licensed files were identified as (near-)duplicates and subsequently removed.

\subsubsection{Capability Enhancement}

LLMs are expected to demonstrate capabilities such as reasoning, mathematical problem-solving, and knowledge memorizing. However, a significant challenge lies in that, in the pre-training process, high-quality capability-related data is sparsely distributed in the entire corpus, and thereby it is difficult for models to be proficient at these above-mentioned capabilities. Previous research, such as work on Qwen~\cite{bai2023qwen}, GLM-130B~\cite{zeng2023glmb}, Nemotron-4~\cite{parmar2024nemotron}, has tried to incorporate instruction-based or high-quality data during the pre-training stage to enhance these abilities. In our study, we collect open-source data from HuggingFace, primarily utilizing the training datasets of various evaluation benchmarks such as MMLU \cite{hendrycks2021measuringmassivemultitasklanguage} and HellaSwag \cite{zellers2019hellaswag}. These data are used experimentally to investigate the relationship between high-quality, capability-focused training data and model performance.

\subsection{Training Configuration}

The pre-training of Moxin 7B spans over 2 trillion tokens and is executed in three distinct phases:
\begin{enumerate}
\item \textbf{Base Pre-training:} In this initial phase, the model is trained on data with a fixed context length of 2,000 tokens. This stage establishes foundational language modeling abilities, capturing syntax and basic semantic patterns.
\item \textbf{Extended Context Pre-training:} In the second phase, the context length is increased to 4,000 tokens, allowing the model to capture longer dependencies and richer context.
\item \textbf{Capability Enhancement:} The final phase incorporates the capability-specific enhancement data described earlier. We provide ablation results that compare the performance with only Phases 1 and 2 versus all three phases, thereby quantifying the benefits of the capability-enhancement data.
\end{enumerate}

We use Colossal-AI \cite{li2023colossal} as our training framework with acceleration techniques including optimized parallelism,  memory optimization, and   heterogeneous training methods.
1) Optimized Parallelism: We combine model, data, and pipeline parallelism to scale training across GPUs. This hybrid parallelism strategy minimizes communication overhead and ensures efficient resource utilization.
2) Memory Optimization: Techniques such as ZeRO redundancy elimination, gradient checkpointing, and mixed-precision training (FP16) are employed to maximize the effective batch size and reduce memory consumption.
3) Heterogeneous Training Methods: the dynamic scheduling and asynchronous communication protocols are leveraged to further boost training throughput. These methods are provided through user-friendly APIs, requiring minimal code modifications.  The training cost is around \$160,000.  

During training, AdamW \cite{loshchilov2017decoupled} with $\beta_1 = 0.9$, $\beta_2 = 0.95$, $\epsilon = 1e^{-8}$ and weight decay = 0.1 is used to optimize the model. We use the cosine learning rate decay, and the learning rate decays to 10\% of its maximum.  
Learning Rate is set to $2e^{-6}$. Our micro-batch size and gradient accumulation steps were tuned to balance between memory constraints and convergence speed.

\section{Model Post-Training}

Following the pre-training phase, we fine-tune the model into a helpful and harmless AI assistant.
We adopt the open-source Tülu 3 dataset and framework \cite{lambert2024t} for our model post-training. Tülu 3 is a family of open state-of-the-art post-trained models, alongside all of the data, data mixes, recipes, code, infrastructure, and evaluation framework, which pushes the boundaries of research in post-training. To close the performance gap between open and closed fine-tuning recipes,  Tülu 3 creates new datasets \& new training procedures, and introduces new methods for training directly on verifiable problems with reinforcement learning. 
The success of  Tülu 3 is rooted in careful data curation, rigorous experimentation, innovative methodologies, and improved training infrastructure, integrating partial details from proprietary methods with novel techniques and established academic research.

For our post-training, with our base model, we follow Tülu 3 to perform supervised finetuning (SFT) and then Direct Preference Optimization (DPO).  Specifically, we use the Tülu 3 SFT Mixture dataset from Tülu 3  to train our base model with  the SFT training method for two epochs and obtain our SFT model, following the default training configuration of the Tülu 3 8B SFT model  \cite{lambert2024t}. To promote  diversity,  the Tülu 3 SFT Mixture dataset contains 939,344 samples from  multiple datasets such as CoCoNot \cite{brahman-kumar2024}, FLAN v2 \cite{weifinetuned}, WildChat GPT-4 \cite{zhao2024wildchat}, etc., together with some data samples collected and released by the  Tülu 3 team.  It  especially consider enhancing several capabilities that can power common use cases and the specific needs, by  including  following datasets: 
OpenMathInstruct \cite{toshniwal2024openmath2} and NuminaMath \cite{numina_math_datasets} for
mathematical reasoning,  
Evol-CodeAlpaca for coding, a subset of Daring-Anteater  \cite{wang2024helpsteer2} for precise instruction following, 
SciRIFF \cite{wadden2024sciriff} for scientific literature understanding, 
and TableGPT \cite{li2023table} for processing table-related tasks.  It  tracks the provenance of dataset and their subsets to verify their licenses
and remove those that have issues. 
With thorough experimentation of Tülu 3, the final SFT data and training hyperparameters are determined to enhance target core skills without significantly impacting the performance
of others, guided by the Tülu 3 evaluation framework. We use the Tülu 3 SFT Mixture dataset  to train our base model with    SFT  for 2 epochs with  a learning rate 5e-06 and a batch size of 128.

Next, we continue to train our SFT  model  on the Tülu 3 8B Preference Mixture dataset from Tülu 3  with the DPO training method to obtain our DPO model, following the same training configuration of the Tülu 3 8B DPO model  \cite{lambert2024t}.    The  data creation pipeline  for the Tülu 3 8B Preference Mixture dataset consists of three stages: prompt selection, response generation from a pool of models, and preference annotation with LLM-as-a-judge to create (preferred, rejected) pairs. The preference mixes data come from different prompt sources, such as SFT data, WildChat and Persona IF. It includes prompts seen during SFT training but also new, unseen prompts. 
To find a better training algorithm and hyper-parameters, Tülu 3 revisits the hyperparameter and algorithm choices alongside the preference datasets. It ablates both algorithm and hyperparameter choices using an early SFT checkpoint and the UltraFeedback dataset.
It explores using DPO, SimPO, and length-normalized DPO, leading to the final training configuration with all hyper-parameters.  We use the Tülu 3 8B Preference Mixture dataset  to finetune  our  SFT model with    DPO   for 1 epochs with  a learning rate 5e-07.

We also adopt the Infinity Instruct dataset \cite{infinityinstruct} to train another version of our instruct model based on our base model.  Infinity Instruct aims to develop a large-scale, high-quality instruction dataset. To construct a ten-million high-quality instruction dataset, it collects a large amount of open-source data as seed and iterate the dataset using two strategies: instruction selection and instruction evolution.
It includes the Foundational Dataset and the Chat Dataset. The Foundational Dataset contains millions of instruction selected from open-source dataset to improve the performance of model on challenging downstream tasks (e.g., code, math). The Chat Dataset contains about 1M instructions evolved from a small subset of high-quality seed data, to further improve the instruction-following ability of model in real conversation scenarios.

\section{Chain-of-Thought Enhancement}

In the context of reasoning capabilities, OpenAI’s o1  series models  firstly introduce inference-time scaling by increasing the length of   Chain-of-Thought (CoT) reasoning process. This approach has achieved significant improvements in various reasoning tasks, such as mathematics, coding, and scientific reasoning. However, the challenge of effective test-time scaling remains an open question for the research community. 
DeepSeek R1 \cite{guo2025deepseek} takes the first step towards improving language model reasoning capabilities using pure reinforcement learning (RL). The goal  is to explore the potential of LLMs to develop reasoning capabilities without any supervised data, focusing on their self-evolution through a pure RL process.  
After thousands of RL steps, DeepSeek-R1-Zero exhibits super performance on reasoning benchmarks.

To enhance the CoT capabilities of our model, we adopt RL techniques similar to DeepSeek R1 \cite{guo2025deepseek}. We first use high quality reasoning data to SFT our instruct model.  The reasoning data mainly includes Openthoughts  \cite{openthoughts}  and OpenR1-Math-220k \cite{openr1math}.  These datasets are obtained by feeding questions to DeepSeek R1 and collecting the reasoning responses from R1. 
Openthoughts  \cite{openthoughts}  contains open synthetic reasoning dataset with 114k high-quality examples covering math, science, code, and puzzles. It generates reasoning traces from DeepSeek-R1 and verifies correctness to construct the final dataset.  
OpenR1-Math-220k \cite{openr1math} is a large-scale dataset for mathematical reasoning. It consists of 220k math problems with two to four reasoning traces generated by DeepSeek R1 for problems from NuminaMath 1.5. 
To build OpenR1-Math-220k, it generates two answers for 400k problems using DeepSeek R1 with 512 H100 GPUs.  Then it applies Math Verify to only retain problems with at least one correct answer, and leverages Llama3.3-70B-Instruct as a judge for 12\% of the samples to retrieve more correct examples, leading to a  filtered dataset containing 220k problems with at least one reasoning trace with a correct answer for each problem.  
We finetune our instruct model on these datasets with SFT based on the open-instruct finetuning framework, following the default training configuration of the Tülu 3 8B SFT model  \cite{lambert2024t},  with    2 epochs  finetuning following   a learning rate 5e-06 and a batch size of 128.

Next after SFT, we further adopt  the RL techniques in DeepSeek R1, i.e., GRPO to  finetune our model with RL.  We adopt the DeepScaleR \cite{deepscaler2025} as our RL training framework.  DeepScaleR is an open-source project to fully democratize RL for LLMs and reproduce DeepSeek R1 and OpenAI O1/O3 at scale on real tasks. It opens source all training scripts (including hyperparameters), models, dataset, and logs. It is based on veRL \cite{sheng2024hybridflow}, Volcano Engine Reinforcement Learning for LLM, a  flexible, efficient and production-ready RL training framework designed for LLMs. 
veRL is flexible and easy to use with 
easy extension of diverse RL algorithms  (combining the strengths of single-controller and multi-controller paradigms to enable flexible representation and efficient execution of complex Post-Training dataflows), 
seamless integration of existing LLM infra with modular APIs (by decoupling computation and data dependencies, and enabling seamless integration with existing LLM frameworks, such as PyTorch FSDP, Megatron-LM and vLLM),
and flexible device mapping (supporting various placement of models onto different sets of GPUs for efficient resource utilization and scalability across different cluster sizes).
veRL is fast with state-of-the-art throughput  and efficient actor model resharding with 3D-HybridEngine. 

DeepScaleR releases DeepScaleR-1.5B-Preview, a 1.5B model, trained on 40K high-quality math problems with 3,800 A100 hours, outperforming OpenAI’s o1-preview on multiple competition-level math benchmarks (such as surpassing O1-Preview and achieves 43.1\% Pass@1 on AIME). It is achieved  by iteratively scaling Deepseek's GRPO algorithm from 8K->16K->24K context length for thinking. 
It releases the training recipe, hyperparameters, data and underlying systems. For the training dataset,  it compiled AIME problems  and AMC problems prior to 2023, along with questions from the Omni-MATH and Still datasets, which feature problems from various national and international math competitions. The data processing pipeline consists of three key steps: extracting answers,  removing redundant questions, and 
filtering ungradable questions.
After deduplication and filtering, the final training dataset consists of approximately 40,000 unique problem-answer pairs.

Different from the original DeepScaleR to train a 1.5B LLM with RL,  we adopt the DeepScaleR  framework to finetune our 7B DPO model with RL for improving CoT capabilities. As advocated in Deepseek-R1, we employ an Outcome Reward Model (ORM) as opposed to a Process Reward Model (PRM) to avoid reward hacking.  The reward function returns: 1 if the LLM’s answer passes basic LaTeX/Sympy checks, and 0 if the LLM’s answer is incorrect or formatted incorrectly.  
We evaluate our COT reasoning model on competition-level mathematics benchmarks, including  AMC 2023, MATH-500, Minerva Math, and OlympiadBench. We report the Pass@1 accuracy. We use Qwen Math's codebase for evaluation following \cite{simplerl,zeng2025simplerl}.

\textcolor{black}{Besides DeepScaleR, we adopt another framework, AReal \cite{mei2024realhf,areal2025} to finetune our  DPO model and obtain another version of our COT model.  AReaL (Ant Reasoning RL) \cite{mei2024realhf,areal2025} is an open-source and efficient reinforcement learning system developed at the RL Lab, Ant Research. AReaL inherits and adapts the Open-Source Project ReaLHF \cite{mei2024realhf} for training Large Reasoning Models (LRMs). Its RL training dataset consists of 40k high-quality mathematical reasoning tasks released by DeepScaleR.  It assigns rewards of +5 for correct responses and -5 for incorrect responses. The correctness of responses is evaluated by comparing the responses with the answers using Sympy, following the evaluation protocol of Qwen. A simplified variant of PPO is adopted as the RL algorithm with a major change to PPO by eliminating the critic to reduce computation resources.  We follow the default configuration of AReal  (with necessary modifications due to our specific environment) to train our 7B DPO model for COT reasoning capabilities. }

\section{Vision Language Model based on Moxin}

Vision  language models (VLMs) \cite{bai2308qwen,zhang2023internlm,zhu2023minigpt,liu2024improved}, which leverage
LLMs, have   achieved significant breakthroughs, enabling sophisticated vision-language dialogues and interactions.  To develop VLM, we adopt the Prismatic VLMs framework \cite{karamcheti2024prismatic} to train our VLM based on our Moxin model. Specifically, for the visual backbone with  image processing  and visual  representations, we adopt DINOv2  \cite{oquab2023dinov2} and SigLIP \cite{zhai2023sigmoid}, and  fuse  their features to  provide significant gains. For the LLM part, we adopt our Moxin-7B-Base as the LLM backbone.

Following the hypotheses in \cite{kerr2023lerf} and similar work, the DINOv2 features provide features that capture low-level spatial properties of an image, augmenting the higher-level “semantic” properties captured by vision-language contrastive models.  
Typically,  the visual module such as (CLIP and SigLIP) trained with the vision-language contrastive objective  leads to much better performance.  Furthermore, SigLIP contains internet-sourced images from multiple sources (e.g., sketches, diagrams, animated graphics, etc.) which are not in ImageNet or in the DINOv2 pretraining data.
DINOv2 and SigLIP  are good complements for the image processing  and visual  representations.

For the model architecture,  we adopt the general architecture used by many recent VLMs, such as LLaVa, Qwen-VL,
and PaLI-3  \cite{liu2023visual,bai2308qwen,chen2023pali}. These architectures use a (pretrained) visual backbone to map an input image to a sequence of patch features that are then projected individually into the embedding
space of an LM.  Specifically,  a VLM takes as input an image and text prompt tokens  with arbitrary sequence length. These inputs are then fed to the following components: 1) a visual representation backbone, 2) a vision-language projector, and 3) a language model. 

We select the pretraining  dataset  that are fully open-source, and have been used in prior works. Specifically, we use the LLaVa v1.5 data mixture, which consists of two subsets used for a multi-stage training pipeline. The first subset consists of a 558K sample mixture of examples sourced from various captioning
datasets (e.g., Conceptual Captions, LAION \cite{sharma2018conceptual}), 
while the second consists of 665K multimodal instruct tuning examples comprised of synthetic data generated in \cite{liu2024mmbench}, as well as
examples from existing vision-language training sets (e.g.,
GQA, TextCaps, \cite{hudson2019gqa}), and notably, a sample of language-only data from
ShareGPT \cite{chen2024sharegpt4v}.

During the training of our VLM,  we adopt a single-stage approach that  both the projection and LM are trained while  the visual representation  modules are  frozen. We train for two epochs.

\section{Evaluation}\label{sec4.evaluation}

We conducted comprehensive performance comparisons against leading language models of comparable scale, including Mistral-7B \cite{jiang2023mistral7b}, LLaMA 2-7B \cite{touvron2023llama}, Gemma-7B \cite{team2024gemma}, and Qwen v2-7B \cite{yang2024qwen2}. These models were selected based on their demonstrated excellence within the 7B or 8B category and represent diverse development approaches from various research organizations worldwide. To ensure a robust evaluation, we re-run all benchmarks with the same evaluation pipeline for fair comparisons. Specifically, we use lm-evaluation-harness \cite{lmevaluationharness}, opencompass \cite{open_compass}, Qwen-Math~\cite{simplerl}, and  OLMES~\cite{lambert2024t} for evaluation.

Lm-evaluation-harness provides a unified framework to test generative language models on a large number of different evaluation tasks. It supports over 60 standard academic benchmarks for LLMs, with hundreds of subtasks and variants implemented.
This framework is versatile as it extends to models implemented through various architectures, including transformers (including quantization via AutoGPTQ~\cite{AutoGPTQ}), GPT-NeoX~\cite{black2022gpt}, and Megatron-DeepSpeed~\cite{song2023deepspeed4science}, all unified through a flexible, tokenization-agnostic interface.
The framework is reliable, as evidenced by serving as the backend for HuggingFace's popular Open LLM Leaderboard and being utilized by dozens of organizations, including NVIDIA, Cohere, BigScience, BigCode, Nous Research, and Mosaic ML. 

To complement, we also employed openCompass. This framework performs an in-depth and holistic assessment of large language models structured around eight fundamental dimensions of language model capabilities: language comprehension, knowledge precision, logical deduction, creative ideation, mathematical problem-solving, programming proficiency, extended text analysis, and intelligent agent engagement.

\subsection{Evaluation Tasks}\label{sec4.1}

We evaluate the model performance on various tasks below.

\begin{itemize}
\item AI2 Reasoning Challenge (ARC)~\cite{allenai:arc} - a set of genuine grade-school level, multiple-choice science questions, assembled to encourage research in advanced question-answering. The dataset is partitioned into a Challenge Set (ARC-C) and an Easy Set (ARC-E), where the former contains only questions answered incorrectly by both a retrieval-based algorithm and a word co-occurrence algorithm.
\item HellaSwag~\cite{zellers2019hellaswag} - a test of commonsense natural language inference, which is easy for humans (~95\%) but challenging for SOTA models. It consists of 70,000 multiple-choice questions. Each question presents a scenario followed by four possible outcomes, asking the model to select the most reasonable conclusion.
\item MMLU~\cite{van2024ai} - a test to measure a text model's multitask accuracy. The test covers 57 tasks, including elementary mathematics, US history, computer science, law, etc.
\item Winogrande~\cite{sakaguchi2021winogrande} - an adversarial and difficult Winograd benchmark at scale, for commonsense reasoning. It contains 44,000 multiple-choice questions with two options each. It requires the model to choose the appropriate entity word for the pronoun in the descriptive text based on the scenario.
\item PIQA~\cite{bisk2019piqareasoningphysicalcommonsense} -   the task of physical commonsense reasoning and a corresponding benchmark dataset Physical Interaction: Question Answering (PIQA). Physical commonsense knowledge is a major challenge on the road to true AI-completeness, including robots that interact with the world and understand natural language. PIQA focuses on everyday situations with a preference for atypical solutions. 
\end{itemize}

\subsection{Pre-Training Evaluation}\label{sec4.2}

We name the initial model as Moxin-7B-Original, which presents the foundation model before fine-tuning on the training data of the evaluation datasets. After subsequent partial fine-tuning of Moxin-7B-Original on the training data of the evaluation datasets, we developed Moxin-7B-Enhanced, enabling direct assessment of how targeted fine-tuning affects model performance.  We release our Moxin-7B-Enhanced  model as Moxin-7B-Base model.

\subsubsection{Zero-Shot Evaluation}

We report the result of base models for zero-shot evaluation in Table \ref{tab:2}. The tasks are listed below. After training with the training data of evaluation tasks, our Moxin-7B-Enhanced can achieve superior performance compared with state-of-the-art (SOTA) baselines. This significant increase from the base model demonstrates the effectiveness of our fine-tuning approach. The improved performance is particularly notable on complex reasoning tasks like PIQA, where the score increased from 78.07\% to 82.24\%, matching or exceeding several leading models.
Consequently, our models emerge as an excellent candidate for real-world applications. 
\begin{itemize}
\item AI2 Reasoning Challenge (0-shot)
\item AI2 Reasoning Easy (0-shot)
\item HellaSwag (0-shot)
\item PIQA (0-shot)
\item Winogrande (0-shot)
\end{itemize}

\scalebox{1.04}{
\begin{threeparttable}[t]
\caption{Performance comparison for various models in zero-shot evaluation.}
\begin{tabular}{c|ccccc|c}
\hline
Models                & HellaSwag & WinoGrade & PIQA  & ARC-E & ARC-C & Ave \\ \hline
Mistral - 7B       & 80.39     & 73.4      & 82.15 & 78.28 & 52.22 & 73.29   \\
LLaMA 2 - 7B          & 75.99     & 69.06     & 79.11 & 74.54 & 46.42 & 69.02   \\
LLaMA 2 - 13B         & 79.37     & 72.22     & 80.52 & 77.4  & 49.06 & 71.71   \\
LLaMA 3.1 - 8B        & 78.92     & 74.19     & 81.12 & 81.06 & 53.67 & 73.79   \\
Gemma - 7b            & 80.45     & 73.72     & 80.9  & 79.97 & 54.1  & 73.83   \\
Qwen v2 - 7B          & 78.9      & 72.38     & 79.98 & 74.71 & 50.09 & 71.21   \\
Internlm2.5 - 7b      & 79.14     & 77.9      & 80.52 & 76.16 & 51.37 & 73.02   \\
Baichuan2 - 7B        & 72.25     & 67.17     & 77.26 & 72.98 & 42.15 & 66.36   \\
Yi-1.5-9B             & 77.86     & 73.01     & 80.74 & 79.04 & 55.03 & 73.14   \\
DeepSeek - 7B         & 76.13     & 69.77     & 79.76 & 71.04 & 44.8  & 68.3    \\ \hline
Moxin - 7B - Original & 72.06     & 66.31     & 78.07 & 71.47 & 48.15 & 67.21   \\
Moxin - 7B - Enhanced & 80.03     & 75.17     & 82.24 & 81.12 & 58.64 & 75.44   \\ \hline
\end{tabular}
\label{tab:2}
\end{threeparttable}}

\subsubsection{Few-Shot Evaluation}

Table \ref{tab:1} presents our zero-shot evaluation results across multiple benchmark tasks. The tasks and their few-show settings are listed below. Thanks to its rigorous and high-quality training corpus, our model demonstrates a remarkable competitive edge in tasks that involve language understanding and knowledge application. Our Moxin-7B-Original achieves superior performance than LLaMA2-7B in this scenario. After training with the training data of evaluation tasks, our Moxin-7B-Enhanced can achieve competitive performance compared with  SOTA baselines.

Consequently, our models emerge as an excellent choice for a multitude of real-world applications where the reliance on robust language comprehension and extensive knowledge is paramount.
\begin{itemize}
\item AI2 Reasoning Challenge (25-shot)
\item HellaSwag (10-shot)
\item MMLU (5-shot)
\item Winogrande (5-shot)
\end{itemize}

\scalebox{1.14}{
\begin{threeparttable}[t]
\caption{Performance comparison for various models in few-shot evaluation.}
\begin{tabular}{c|cccc|c}
\hline
Model                  & ARC-C   & Hellaswag & MMLU  & WinoGrade & Ave   \\ \hline
Mistral - 7B      & 57.59 & 83.25     & 62.42 & 78.77     & 70.51 \\
LLaMA 3.1 - 8B         & 54.61 & 81.95     & 65.16 & 77.35     & 69.77 \\
LLaMA 3 - 8B           & 55.46 & 82.09     & 65.29 & 77.82     & 70.17 \\
LLaMA 2 - 7B           & 49.74 & 78.94     & 45.89 & 74.27     & 62.21 \\
Qwen 2 - 7B            & 57.68 & 80.76     & 70.42 & 77.43     & 71.57 \\
Gemma - 7B             & 56.48 & 82.31     & 63.02 & 78.3      & 70.03 \\
Internlm2.5 - 7B       & 54.78 & 79.7      & 68.17 & 80.9      & 70.89 \\
Baichuan2 - 7B         & 47.87 & 73.89     & 54.13 & 70.8      & 61.67 \\
Yi-1.5-9B              & 58.36 & 80.36     & 69.54 & 77.53     & 71.48 \\ \hline
Moxin - 7B - Original  & 53.75 & 75.46     & 59.43 & 70.32     & 64.74 \\
Moxin - 7B - Enhanced & 59.47 & 83.08     & 60.97 & 78.69     & 70.55 \\ \hline
\end{tabular}
\label{tab:1}
\end{threeparttable}}

\subsection{Post-Training Evaluation}  

In our post-training, with Tülu 3,  we further fine-tune our base model with SFT and then DPO,  leading to our Moxin-7B-SFT model and Moxin-7B-DPO model, respectively. We also adopt the Infinity Instruct dataset \cite{infinityinstruct} to train another version of our instruct model based on our base model, resulting in Moxin-7B-DPO-II.
We demonstrate the performance in  Table~\ref{tab:post-zero-shot} for zero-shot evaluation and Table~\ref{tab:post-few-shot} for few show evaluation. We also use the OLMES framework from Tulu 3 \cite{lambert2024t} to evaluate our model across a range of tasks and show the results in Table \ref{tab:post-olmes}.   We can observe that our Moxin-7B-DPO model can achieve comparable performance with other SOTA instruct models.  We release the  Moxin-7B-DPO  model as Moxin-7B-Instruct model.

\scalebox{1.1}{
\begin{threeparttable}[t]
\caption{Performance comparison for various models in zero-shot evaluation.}
\begin{tabular}{c|ccccc|c}
\hline
Models               & HellaSwag & WinoGrade & PIQA  & ARC-E & ARC-C & Ave   \\ \hline
Mistral 8B Instruct  & 79.08     & 73.56     & 82.26 & 79.88 & 56.57 & 74.27 \\
Llama3.1 8B Instruct & 79.21     & 74.19     & 80.79 & 79.71 & 55.03 & 73.79 \\
Qwen2.5 7B Instruct  & 80.5      & 71.03     & 80.47 & 81.31 & 55.12 & 73.69 \\ \hline
Moxin - 7B - II     & 79.32	& 72.93	& 81.56 &	80.43 &	56.91	 & 74.23\\
Moxin - 7B - SFT     & 81.44     & 73.09     & 81.07 & 79.8  & 54.67 & 74.01 \\
Moxin - 7B - DPO     & 85.7      & 73.24     & 81.56 & 81.1  & 58.02 & 75.92 \\ \hline
\end{tabular}
\label{tab:post-zero-shot}
\end{threeparttable}}

\scalebox{1.2}{
\begin{threeparttable}[t]
\caption{Performance comparison for various models in few-shot evaluation.}
\begin{tabular}{c|cccc|c}
\hline
Model                & ARC-C & Hellaswag & MMLU  & WinoGrade & Ave  \\ \hline
Mistral 8B Instruct  & 62.63 & 80.61     & 64.16 & 79.08     & 71.62    \\
Llama3.1 8B Instruct & 60.32 & 80        & 68.18 & 77.27     & 71.44    \\
Qwen2.5 7B Instruct  & 66.72 & 81.54     & 71.3  & 74.59     & 73.54    \\ \hline
Moxin - 7B - II     & 61.35& 	82.1	& 62.95	& 77.98	&  71.095  \\
Moxin - 7B - SFT     & 60.11 & 83.43     & 60.56 & 77.56     & 70.42    \\
Moxin - 7B - DPO     & 64.76 & 87.19     & 58.36 & 76.32     & 71.66    \\  \hline
\end{tabular}
\label{tab:post-few-shot}
\end{threeparttable}}

\scalebox{0.78}{
\begin{threeparttable}[t]
\caption{Performance comparison for various models in olmes evaluation.}
\begin{tabular}{c|cccccccc|c}
\hline
Models/Datasets      & GSM8K & MATH  & Humaneval & \begin{tabular}[c]{@{}c@{}}Humaneval\\ plus\end{tabular} & MMLU  & PopQA & BBH   & TruthfulQA & Ave   \\ \hline
Qwen2.5 7B Instruct  & 83.8  & 14.8  & 93.1      & 89.7                                                     & 76.6  & 18.1  & 21.7  & 63.1       & 57.61 \\
Gemma2 9B Instruct   & 79.7  & 29.8  & 71.7      & 67                                                       & 74.6  & 28.3  & 2.5   & 61.4       & 51.88 \\
\hline
Moxin - 7B - II  & 71.04 &  21   & 78.21  & 72.35  & 63.27  &  27.98  & 44.33 &   56.22 &  54.42 \\
Moxin - 7B - DPO         & 81.19 & 36.42 & 82.86     & 77.18                                                    & 60.85 & 23.85 & 57.44 & 55.27      & 59.38 \\ \hline
\end{tabular}
\label{tab:post-olmes}
\end{threeparttable}}


\subsection{CoT Evaluation}

Based on our instruct model, we further finetune the model with reasoning data based on SFT and then train  using reinforcement learning with GRPO following DeepSeek R1.   
Starting from  the same model after SFT with the OpenThoughts and Open-R1-Math-220k datasets, we use two different RL frameworks, i.e., DeepScaleR \cite{deepscaler2025}  and AReal \cite{areal2025}, to develop two versions of our reasoning model, respectively,  resulting in Moxin-7B-RL-DeepScaleR and   Moxin-7B-RL-AReal. 
We evaluate our  reasoning models on competition-level mathematics benchmarks, including  AMC 2023, MATH-500, Minerva Math, and OlympiadBench. We report the Pass@1 accuracy. We use Qwen Math's codebase for evaluation following \cite{simplerl,zeng2025simplerl} and show the comparison in Table~\ref{tab:cot}.   We can observe that the Moxin-7B-RL-DeepScaleR model can achieve outstanding performance compared with baselines,  demonstrating the effectiveness of RL for small LLMs such as 7B models.  Our  Reasoning model trained with DeepScaleR performs better than that trained with AReal. Our reasoning model Moxin-7B-RL-DeepScaleR trained with DeepScaleR \cite{deepscaler2025} is  released as  Moxin-7B-Reasoning model.

\scalebox{0.86}{
\begin{threeparttable}[t]
\caption{Performance comparison for various models on reasoning evaluation.}
\begin{tabular}{c|cccc|c}
\hline
Models/Datasets                    & MATH 500 & AMC  & Minerva Math & OlympiadBench & Ave   \\ \hline
Qwen2.5-Math-7B-Base               & 52.4     & 52.5 & 12.9         & 16.4          & 33.55 \\
Qwen2.5-Math-7B-Base + 8K MATH SFT & 54.6     & 22.5 & 32.7         & 19.6          & 32.35 \\
Llama-3.1-70B-Instruct             & 64.6     & 30.1 & 35.3         & 31.9          & 40.48 \\ \hline
Moxin-7B-RL-AReal               & 68.6       & 50 & 16.9         & 31.7          & 41.8  \\
Moxin-7B-RL-DeepScaleR                & 68       & 57.5 & 16.9         & 30.4          & 43.2  \\ \hline
\end{tabular}
\label{tab:cot}
\end{threeparttable}}

\subsection{VLM Evaluation}

We adopt the evaluation suit in Prismatic VLMs \cite{karamcheti2024prismatic}  to evaluate the VLM performance. It focuses on evaluations with well-defined metrics, spanning the following three areas: 

\textit{Open-Ended Visual Question Answering.} 
We evaluate on VizWiz \cite{bigham2010vizwiz} and GQA \cite{hudson2019gqa}.  VizWiz assess general visual reasoning; VizWiz also contains a series of unanswerable questions. GQA evaluates spatial reasoning.

\textit{Localization.} Part of the pretraining data mixture  contains examples of predicting normalized bounding box coordinates given referring expressions in language. As such, we evaluate bounding box prediction accuracy on RefCOCO, RefCOCO+, and RefCOCOg \cite{kazemzadeh2014referitgame}, and on OCID-Ref \cite{wang2021ocid}. RefCOCO focuses on short descriptions with spatial anchors, RefCOCO+ on strictly appearance based descriptions, and RefCOCOg on long, rich descriptions; OCID-Ref is a robotics dataset probing out-of-distribution generalization, with a focus on localizing objects in clutter.

\textit{Challenge Sets (Closed-Set Prediction).}  We evaluate on Visual Spatial Reasoning \cite{liu2023visual}, TallyQA \cite{acharya2019tallyqa}, and POPE \cite{liu2023query}. VSR consists of challenging True/False questions about individual spatial relationships in diverse scenes; this is an especially challenging task, with most existing models failing to outperform the majority class baseline. TallyQA consists of questions that assess a VLM’s ability to count objects described in language, with expressions that range in complexity. POPE consists of targeted Yes/No questions that assess a VLM’s propensity to hallucinate. 

We use the validation sets for all benchmarks except GQA (where use the recommended the test-dev split), VSR (where we use the zero-shot test split), and POPE (where there is only a single evaluation split).

The comparisons with other VLMs are demonstrated in Table \ref{tab:vlm}. We compare with LLaVa v1.5 7B. Furthermore, we adopt the same VLM framework and change the LLM backbone to other LLMs such as Llama2 7B and Mistral 7B. Then we train these VLMs and compare with our VLM based on Moxin-7B. We can observe that our model outperforms all other VLM baselines. 

\scalebox{0.86}{
\begin{threeparttable}[t]
\caption{Performance comparison for various VLM models.}
\begin{tabular}{l|ccccccc|c}
\hline
                         & GQA   & VizWiz & RefCOCO+ & OCID-Ref & VSR   & POPE  & TallyQA & Ave.  \\ \hline
LLaVa v1.5 7B (Base)     & 61.58 & 54.25  & 49.47    & 35.07    & 51.47 & 86.57 & 62.06   & 57.21 \\
Llama-2 Chat 7B          & 62.11 & 56.39  & 58.5     & 46.3     & 61.8  & 86.8  & 58.1    & 61.43 \\
Mistral v0.1 7B          & 63.3  & 55.32  & 65.1     & 48.8     & 58.5  & 87.1  & 61.7    & 62.83 \\
Mistral Instruct v0.1 7B & 62.71 & 54.35  & 64.9     & 48       & 57.8  & 87.5  & 64.5    & 62.82 \\
Llama-2 7B               & 62.44 & 55.98  & 59.47    & 43.89    & 63.67 & 86.74 & 59.22   & 61.63 \\ \hline
Ours                     & 64.88 & 54.08  & 71.3     & 48.4     & 60.8  & 87.3  & 66      & 64.68 \\ \hline
\end{tabular}
\label{tab:vlm}
\end{threeparttable}}

\subsection{Moxin Family} 

We demonstrate a list of our models in the paper in Table~\ref{tab:models}. Multiple models are developed in our paper, and we show their names in our releases.

\begin{center}
\scalebox{1.2}{
\begin{threeparttable}[h]
\caption{Our developed models and their names in our releases.}
\begin{tabular}{l|l}
\hline
Developed Models       & Names in Releases  \\ \hline
Moxin-7B-Enhanced     & Moxin-7B-LLM     \\ \hline
Moxin-7B-SFT           &                    \\\hline
Moxin-7B-DPO           & Moxin-7B-Instruct  \\\hline
Moxin-7B-DPO-II        &                    \\ \hline
Moxin-7B-RL-DeepScaleR & Moxin-7B-Reasoning \\\hline
Moxin-7B-RL-AReal      &                    \\ \hline
Moxin-7B-VLM      &  Moxin-7B-VLM                \\ \hline
\end{tabular}
\label{tab:models}
\end{threeparttable}}
\end{center}

\subsection{Generation Examples}  
We demonstrate some text generation examples from our Moxin Chat 7B and Llama2 Chat 7B in Table \ref{tab:E1} and \ref{tab:E2}. Moxin Chat 7B is developed by finetune the Moxin base model on the Tulu2 dataset. 
The response of Moxin Chat 7B is more accurate and concise.

\begin{center}
\begin{threeparttable}[]

\scalebox{0.7}{
\begin{tabular}{l|l}
\hline
               & Example 1                                                                                                                                                                                                                                                                                                                                                                                                                                                                                                                                                                                                                                                                                                                                                                                                                                                                                                                                                                                                                                                                                                                                                                                                                                                                                                                                                                                                                                                                                                                                                                                                                                                                                                                                                                                                                                                                                                                                                                                                                                                                                                                                                                                                                                                                                                                                                                                                                                                                                                                                                                                                                                                                                                                                                                                                                                                                                                                                                                                                                                                                                                                                                                                                                                                                                                                    \\ \hline
Moxin Chat 7B  & \begin{tabular}[c]{@{}l@{}}Can you explain the concept of regularization in machine learning? \\ \\ Regularization is a technique used in machine learning to prevent overfitting by introducing a penalty \\ term in the loss function. The idea behind regularization is to encourage simpler models that are easier to \\ interpret and generalize better to new data. \\ \\ There are two main types of regularization: \\ 1. L1 regularization: also known as Lasso regularization, this method adds a penalty term to the loss \\ function that is proportional to the absolute value of the coefficients in the model. This encourages \\ sparsity, meaning that some coefficients may be exactly zero, which simplifies the model and makes it \\ easier to interpret. \\ 2. L2 regularization: also known as Ridge regularization, this method adds a penalty term to the loss \\ function that is proportional to the square of the coefficients in the model. This encourages shrinkage, \\ meaning that the coefficients are pulled closer to zero, which simplifies the model and reduces the \\ variance of the model estimates. \\ \\ Regularization can be used in various machine learning models, including linear regression, logistic \\ regression, and neural networks. It is often used in conjunction with other techniques such as cross-\\ validation to find the optimal hyperparameters that balance the bias-variance tradeoff and prevent \\ overfitting.\end{tabular}                                                                                                                                                                                                                                                                                                                                                                                                                                                                                                                                                                                                                                                                                                                                                                                                                                                                                                                                                                                                                                                                                                                                                                                                                                                                                                                                                                                                                                                                                                                                                                                                                                                                                                                                                                                                                         \\ \hline
Llama2 Chat 7B & \begin{tabular}[c]{@{}l@{}}Can you explain the concept of regularization in machine learning? \\ \\ Regularization is a technique used in machine learning to prevent overfitting. It involves adding a \\ penalty term to the loss function to discourage large values of the model's weights. \\ \\ There are several types of regularization, including: \\ 1. L1 regularization (Lasso): This adds a penalty term to the loss function that is proportional to the \\ absolute value of the model's weights. This encourages the model to have smaller weights, which can \\ help prevent overfitting. \\ 2. L2 regularization (Ridge): This adds a penalty term to the loss function that is proportional to the \\ square of the model's weights. This also encourages the model to have smaller weights, but it can lead \\ to some weights being set to zero, which can result in a simpler model. \\ 3. Dropout regularization: This is a type of regularization that is applied during training. It involves \\ randomly setting a fraction of the model's neurons to zero during each training iteration. This helps to \\ prevent overfitting by making the model less reliant on any single neuron. 4. Early stopping: This is a \\ type of regularization that is applied during training. It involves monitoring the validation loss during \\ training and stopping the training process when the validation loss stops improving. This helps to prevent \\ overfitting by stopping the training process before the model has a chance to fit the noise in the training \\ data. \\ \\ Regularization can be applied to different layers of a neural network, such as the input layer, hidden \\ layers, or output layer. It can also be applied to different parts of a model, such as the weights, biases, or \\ activation functions. The main idea behind regularization is to add a penalty term to the loss function that \\ discourages large values of the model's weights. This helps to prevent overfitting by encouraging the \\ model to find a simpler solution that generalizes better to new data. \\ \\ Regularization can be used in conjunction with other techniques, such as cross-validation and grid search, \\ to improve the performance of a machine learning model. In summary, regularization is a technique used \\ in machine learning to prevent overfitting by adding a penalty term to the loss function. There are several \\ types of regularization, including L1, L2, dropout, and early stopping, and it can be applied to different \\ layers and parts of a model. Regularization helps to find a simpler solution that generalizes better to new \\ data and can be used in conjunction with other techniques to improve the performance of a machine \\ learning model.\end{tabular} \\ \hline
\end{tabular}}
\caption{Generation Examples.} \label{tab:E1}
\end{threeparttable}
\end{center}

\begin{center}
\begin{threeparttable}[t]
\scalebox{0.7}{
\begin{tabular}{l|l}
\hline                                                               & Example 2                                                                                                                                                                                                                                                                                                                                                                                                                                                                                                                                                                                                                                                                                                                                                                                                                                                                                                                                                                                                                                                                                                                                                                                                                                                                                                                                                                                                                                                                                                                                                                                                                                                                                                                                                                                                                                                                                                                                                                                                                                                                                                                                                                                                                                                                                                                                                                                                                                                                                                                                                                                                                                                                                                                                                                                                                                                                                                                                                                                                                                                                                                                                                                                                                                                                                                                    \\ \hline
Moxin Chat 7B  & \begin{tabular}[c]{@{}l@{}}Can you explain the concept of normalization in machine learning? \\ \\ Normalization is a technique used to scale the data before using it for training a machine learning model. \\ It is used to make the data more uniform, and to prevent it from being skewed by large or small values. \\ \\ There are several ways to normalize data, including min-max scaling, z-score normalization, and \\ logarithmic scaling. In min-max scaling, the values are scaled to a range between 0 and 1, by dividing \\ them by the maximum value in the dataset. In z-score normalization, the values are scaled to have a mean \\ of 0 and a standard deviation of 1, by subtracting the mean from each value and dividing by the standard \\ deviation. In logarithmic scaling, the values are transformed using the logarithm function, which can be \\ useful for normalizing data with a wide range of values. \\ \\ Normalization is an important step in preparing data for machine learning, as it can help to improve the \\ performance of the model by making the data more uniform and easier to analyze. It can also help to \\ prevent overfitting and improve generalization.\end{tabular}                                                                                                                                                                                                                                                                                                                                                                                                                                                                                                                                                                                                                                                                                                                                                                                                                                                                                                                                                                                                                                                                                                                                                                                                                                                                                                                                                                                                                                                                                                                                                                                                                                                                                                                                                                                                                                                                                                                                                                                                                                                                                               \\ \hline
Llama2 Chat 7B & \begin{tabular}[c]{@{}l@{}}Can you explain the concept of normalization in machine learning? \\ \\ Normalization is a technique used in machine learning to transform input data into a range of values that \\ are suitable for use in a model. The goal of normalization is to ensure that the data is on the same scale, so \\ that the model can learn meaningful relationships between the features. \\ \\ There are several types of normalization techniques, including: \\ 1. Min-max normalization: This involves scaling the data so that it falls within a specific range, usually \\ between 0 and 1. \\ 2. Standardization: This involves subtracting the mean and dividing by the standard deviation of the data, \\ so that the data has a mean of 0 and a standard deviation of 1. \\ 3. Z-score normalization: This involves subtracting the mean and dividing by the standard deviation of \\ the data, so that the data has a mean of 0 and a standard deviation of 1. \\ 4. Log transformation: This involves taking the logarithm of the data, which can help to reduce the effect \\ of extreme values in the data. \\ 5. Square root transformation: This involves taking the square root of the data, which can help to reduce \\ the effect of outliers in the data. \\ \\ Normalization is an important preprocessing step in machine learning, as it can help to improve the \\ performance of the model by reducing the effect of irrelevant features and improving the interpretability \\ of the results. \\ \\ Here are some key points to keep in mind when normalizing data for machine learning: \\ 1. Normalization can improve the performance of the model by reducing the effect of irrelevant features \\ and improving the interpretability of the results. \\ 2. There are several types of normalization techniques, including min-max normalization, standardization, \\ z-score normalization, log transformation, and square root transformation. \\ 3. The choice of normalization technique will depend on the nature of the data and the specific \\ requirements of the model. \\ 4. Normalization can be applied to both continuous and categorical data. \\ 5. It is important to normalize the data before training the model, as normalization can improve the \\ convergence of the model and reduce the risk of overfitting. \\ 6. Normalization can also be used to improve the interpretability of the results by reducing the effect of \\ irrelevant features and improving the scalability of the model. \\ 7. It is important to carefully evaluate the performance of the normalized data and consider the trade-offs \\ between different normalization techniques. \\ 8. Normalization can be applied to different features of the data, such as the input features, the target \\ variable, or the output features. \\ 9. Normalization can be used in combination with other preprocessing techniques, such as feature scaling, \\ feature selection, and data augmentation. \\ 10. Normalization is a powerful tool for improving the performance of machine learning models, but it is \\ important to carefully evaluate the performance of the normalized data and consider the trade-offs \\ between different normalization techniques.\end{tabular} \\ \hline
\end{tabular}}
\caption{Generation Examples.}
\label{tab:E2}
\end{threeparttable}
\end{center}

\section{Conclusion}\label{sec5.conclusion}

The field of Large Language Models has witnessed a significant shift toward open-source development, fostering innovation within the AI community. However, a critical challenge emerges: many purportedly open-source models withhold essential components necessary for full understanding and reproducibility, creating barriers that limit both academic advancement and commercial adoption. This not only hampers scientific progress, but also prevents businesses from fully leveraging these models for innovative applications, ultimately diminishing potential societal benefits and economic value creation.
To address these limitations, we introduce Moxin 7B, a fully open-source language model developed in accordance with the Model Openness Framework (MOF), providing comprehensive access to pre-training code, configurations, training and fine-tuning datasets, and all intermediate checkpoints. 
We further finetune the Moxin Base model with open-source post-training frameworks  and instruction/CoT data to obtain Moxin Instruct  and Moxin Reasoning models. 
Experiments show that our models achieve superior performance in various evaluations such as zero-shot, few-shot, and CoT evaluations. 
We wish to see more work that establishes new standards for reproducible research in language model development, fostering a more inclusive and economically vibrant AI ecosystem.



\bibliographystyle{unsrt}
\bibliography{sample}

\end{document}